\documentclass[preprint,11pt]{elsarticle}
\usepackage[utf8]{inputenc}
\usepackage{CJK}
\usepackage{amssymb}
\usepackage{autobreak}
\usepackage{multirow}
\usepackage{threeparttable}
\biboptions{numbers,sort&compress}
\usepackage{subfigure}
\allowdisplaybreaks[4]
\usepackage{geometry}
\usepackage{xcolor}
\usepackage{enumitem}
\usepackage{amsmath}
\usepackage{booktabs}
\usepackage{longtable}
\usepackage{rotating}
\usepackage{lscape}
\usepackage{array}
\usepackage{adjustbox}
\usepackage{graphicx}
\usepackage{acronym}
\usepackage{float}
\usepackage[final]{changes}
\usepackage{tabularx}  
\usepackage{booktabs}  
\usepackage{hyperref}
\usepackage{xurl}
\usepackage{nomencl}  
\usepackage{nomencl}  
\usepackage{ifthen}  
\makenomenclature  

\journal{Applied Computing}

\begin{document}

\begin{frontmatter}


\title{LLM-YOLOMS: Large Language Model-based Semantic Interpretation and Fault Diagnosis for Wind Turbine Components}


\author{Yaru Li$^{a}$\corref{}}
\author{Yanxue Wang$^{a}$\corref{cor1}}
\cortext[cor1]{Corresponding author Tel.: +86 15910596997}
\ead{yan.xue.wang@gmail.com}
\author{Meng Li$^{a}$}
\author{Xinming Li$^{a}$}
\author{Jianbo Feng$^{a}$}

\address{
    $^a$School of Mechanical-Electronic and Vehicle Engineering, Beijing University of Civil Engineering and Architecture, Beijing 100044, China\\
    }

\begin{abstract}
The health condition of wind turbine (WT) components is crucial for ensuring stable and reliable operation. However, existing fault detection methods are largely limited to visual recognition, producing structured outputs that lack semantic interpretability and fail to support maintenance decision-making. To address these limitations, this study proposes an integrated framework that combines YOLOMS with a large language model (LLM) for intelligent fault analysis and diagnosis. Specifically, YOLOMS employs multi-scale detection and sliding-window cropping to enhance fault feature extraction, while a lightweight key-value (KV) mapping module bridges the gap between visual outputs and textual inputs. This module converts YOLOMS detection results into structured textual representations enriched with both qualitative and quantitative attributes. A domain-tuned LLM then performs semantic reasoning to generate interpretable fault analyses and maintenance recommendations. Experiments on real-world datasets demonstrate that the proposed framework achieves a fault detection accuracy of 90.6\% and generates maintenance reports with an average accuracy of 89\%, thereby improving the interpretability of diagnostic results and providing practical decision support for the operation and maintenance of wind turbines.

\end{abstract}




\end{frontmatter}

\section{Introduction}

Wind energy is a clean and renewable resource. In recent years, it has been extensively developed and widely adopted worldwide \cite{he2023optimal,zhu2024optimizing,narayanan2025adaptive}. According to the 2024 report by the World Wind Energy Association, the global installed capacity of Wind Turbine (WT) reached 1,097 gigawatts as of June. In the same year, a series of blade failures at a wind power plant in Sweden resulted in substantial losses and the suspension of operations for that particular WT model, causing Siemens Energy a cumulative financial loss of €4.5 billion. Blades are critical components of the WT system and play an essential role in maintaining overall operational stability \cite{song2025optimal,lu2020india}. Damage to wind turbine components such as blades significantly reduces power generation efficiency and may lead to the collapse of the entire turbine, resulting in serious economic losses and safety hazards \cite{fan2024dewp,guo2021nacelle,long2023role}. As a result, monitoring the health status of WT components has become a major research focus, attracting considerable attention from both academia and industry \cite{yimam2025state,mourad2023failure}.

WT components inspection methods have undergone a significant evolution from traditional techniques to modern methods. Early manual visual inspections are inefficient and difficult to detect internal damage, while sensor-based monitoring had problems with complex wiring and signal delays. Vibration-based detection methods suffered from low accuracy due to limited sensor precision. SCADA-based fault detection methods are constrained by weak data processing capabilities\cite{fan2024multiscale}. With the development of computer vision, Unmanned Aerial Vehicles (UAVs)-based inspection systems have become widely used. These systems collect high-resolution, multi-angle images that provide essential data for target detection algorithms \cite{ashkarkalaei2025feature,ye2024wind,hu2025digital}. In the field of target detection, traditional two-stage methods offer high accuracy. However, their complex computations and slower speed have led to the rise of single-stage methods \cite{gohar2025review,ren2015faster}. Single-stage methods complete detection in a single forward pass. This efficiency makes them more suitable for real-time WT component inspections. As a result, they have become the mainstream approach for WT component fault detection \cite{tan2025multi}.

As a representative of single-stage detection, YOLO performs well in the field of WT component detection. With its powerful feature learning and target positioning capabilities, it can quickly and efficiently detect various fault-related targets in images \cite{wang2025precision,akdougan2025pp,bakirci2024enhancing,zhou2024yolo}. Many researchers have focused on optimizing YOLO and exploring its applications\cite{qi2025yolo,hou2025mfel,wang2025precision}. Zhang et al. \cite{zhang2024wind} used synthetic datasets and genetic algorithms to fine-tune YOLOv5. This approach improved both accuracy and robustness. Li et al. \cite{li2025large} proposed an enhanced YOLO model by integrating a squeeze-and-excitation attention mechanism. Their improvements significantly increased training efficiency and detection performance. Zhao et al. \cite{zhao2025wind} developed an image fusion algorithm and constructed the RBs-YOLO model for rotating component monitoring. Their model achieved promising results. Despite these advancements, challenges remain. WT environments are complex, faults are rare, and collecting high-quality fault samples is expensive \cite{gao2026coupled}. As a result, available training data is limited. This shortage affects the accuracy of YOLO models and their ability to detect complex faults. To address this issue, this study proposes a new model called YOLOMS. It uses multi-scale cropping and sliding window techniques for data augmentation. These methods help extract more information from limited samples. YOLO is then used to perform fault detection and diagnosis. This framework improves detection performance in data-scarce conditions.

Despite the continuous advancements in YOLO-based detection frameworks, including YOLOMS, existing models still face significant limitations in their output paradigm, which remains overly standardized and exhibits limited interpretability\cite{cai2025systematic,stodt2023study,ali2024yolo}. The core output of YOLOMS is restricted to structured data consisting of bounding box coordinates, fault category labels, and confidence scores\cite{li2025oam}. This format does not support causal reasoning regarding the mechanisms of fault generation and fails to establish a connection between detection outcomes and the operational and maintenance requirements of wind turbines\cite{truong2023towards,murat2025comprehensive}. Such a result-oriented pattern, lacking causal logic, forces researchers to perform additional domain-specific interpretation of detection data, thereby increasing the complexity and uncertainty of subsequent fault analysis. Furthermore, YOLOMS provides no capability for maintenance decision support; it cannot deliver targeted diagnostic insights or generate engineering-oriented maintenance recommendations. As a result, its functionality remains confined to fault localization, making it difficult to support downstream diagnostic reasoning or maintenance execution workflows, ultimately creating a technological gap between detection and diagnosis as well as between analytical outputs and decision-making\cite{yang2023survey}.

In recent years, Large Language Models (LLMs) as a frontier technology in artificial intelligence, it have emerged as a key enabler for overcoming these limitations. Built upon semantic networks containing hundreds of billions of parameters, LLMs provide a powerful framework for multimodal interaction and have demonstrated remarkable logical reasoning capabilities across diverse domains, including medical diagnostics and financial risk management \cite{pele2026beginning,liu2025query,kung2023performance,adeshola2024opportunities,zheng2024empirical,alonso2023analysis}. Unlike YOLOMS’s single-output paradigm, LLMs exhibit significant advantages in interpretability, interactive analysis and maintenance recommendation generation\cite{zhang5313705treeqa,zhang2025llm,pendyala2025performance}.

From the interpretability perspective, LLMs integrate domain knowledge to convert YOLOMS’s structured detection outputs into natural language explanations with clear causal logic, describing the basic characteristics of faults while providing causal inference, risk assessment, and other key analytical dimensions. This substantially reduces researchers’ cognitive load and interpretive cost. From the interactive analysis perspective, LLMs support context-aware\cite{wang2025ptfusion}, dialogue-based interactions\cite{xiao2025leveraging}, enabling researchers to iteratively probe deeper into initial detection results. The model can synthesize prior outputs with domain expertise to provide coherent and targeted responses\cite{cui2025integrating,xu2025staf}, thereby expanding the depth and continuity of fault analysis and overcoming the one-shot limitation of YOLO-style outputs. In maintenance recommendation generation, LLMs can incorporate wind power maintenance standards and real-world engineering experience to produce actionable maintenance strategies and execution suggestions—based on fault type, severity, and turbine operational conditions. This capability effectively bridges the gap between fault detection and O\&M decision-making, addressing a long-standing technical deficiency in conventional object detection models.

A modality gap exists between visual detection outputs and LLM textual inputs, often leading to the loss of critical information. Moreover, general-purpose LLMs show limited adaptation to the domain-specific knowledge required for wind turbine applications\cite{liu2025cpllm,wang2025relative}. To address these challenges, this study tackles two key issues: introducing a lightweight key-value (KV) mapping mechanism to bridge modalities by transforming visual features into structured textual representations without information loss, and constructing a high-quality domain-specific fine-tuning dataset from expert maintenance logs to enhance the LLM’s professional reasoning capability. This enables end-to-end intelligent inference across fault identification, mechanism analysis, and maintenance decision-making, forming a closed-loop solution for detection, diagnosis, and maintenance. The contributions of this work are as follows:

(1) We propose LLM-YOLOMS, a framework that integrates YOLOMS-based fault feature extraction with the domain reasoning capabilities of fine-tuned LLMs to achieve comprehensive and accurate fault detection, diagnosis, and maintenance recommendations for wind turbine components.

(2) A lightweight cross-modal transformation KV method is proposed, which extracts multi-dimensional vectors from YOLOMS and converts them into structured text suitable for LLM; it calculates fault frequency and damage area to generate qualitative and quantitative descriptions. This method, as a core bridge between the two, prevents information loss due to label overlap and provides preliminary semantic input for subsequent fault analysis.

(3) This paper proposes a high-precision YOLOMS model that uses multi-scale cropping and sliding window cropping methods, which enriches the number of samples in the dataset and helps in the extraction of fault features. Experimental results show that YOLOMS has better detection capabilities than other versions of the YOLO series.

(4) To address the lack of WT component domain expertise in LLM, this study constructs a question-answering dataset using maintenance logs and domain knowledge. The dataset is used to fine-tune the LLM, enabling it to learn WT component fault detection and provide professional, comprehensive support for O\&M.

(5) The framework proposed in this paper was tested on a validation set. The results show that it has significant advantages in fault detection accuracy, semantic understanding capability, and maintenance decision support.

This work proposes a fault detection and diagnosis method for WT components based on the integration of YOLOMS and LLM. Section 2 introduces the computational process of the proposed method. Section 3 presents the experimental results and analysis based on real-world WT component data. The last section concludes the study.

\section{Health monitoring method for target structures in the working state of WT components}

\begin{figure}[htbp]
    \centering
    \includegraphics[width=1\linewidth]{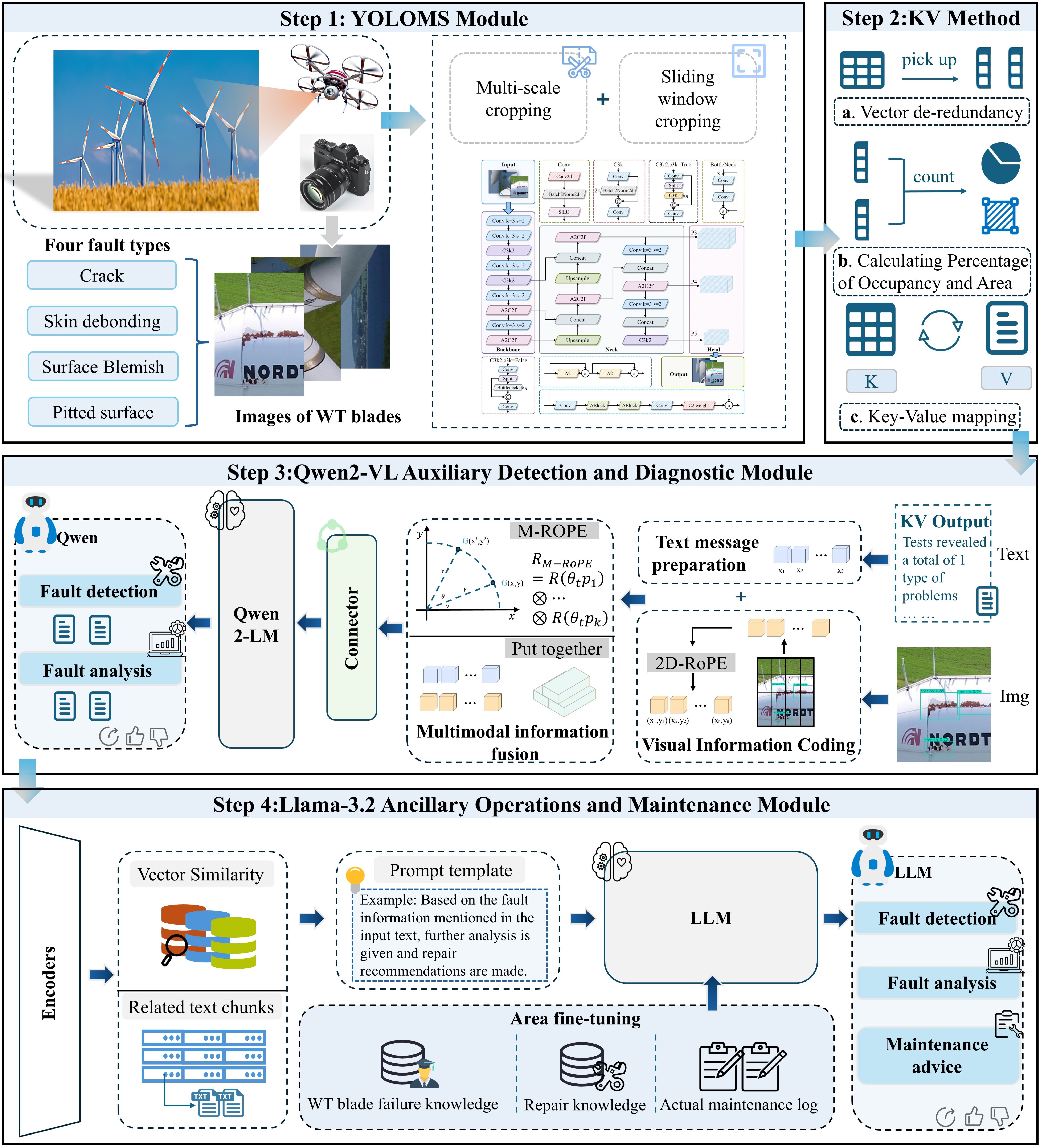}
    \caption{Diagram of the framework using YOLOMS for LLM-assisted wind turbine component fault detection and diagnosis.}
    \label{fig:1}
\end{figure}

This study proposes a fault detection and diagnosis method for WT components based on YOLOMS-Qwen-Llama, utilizing image data collected by UAVs. The network framework of the proposed method is shown in Fig.\ref{fig:1}.

The proposed method uses a multi-stage process to detect component faults. The steps are as follows:

Step 1: High-definition images of WT components taken by UAVs are fed into the YOLOMS module for multi-scale and sliding window cropping. This increases the dataset size and improves feature extraction efficiency. The expanded dataset is then input into the YOLOv12 detection model, which outputs images with recognition frames and corresponding feature vectors.

Step 2: These feature vectors are transformed into concise textual descriptions using a lightweight Key-Value (KV) image-text mapping approach, which incorporates frequency and area analysis of each fault type.

Step 3: The text from the KV method and the images with recognition frames are sent to the Qwen2-VL module for auxiliary fault detection and diagnosis. This step combines multimodal information to produce comprehensive detection results and fault analysis.

Step 4: The output from Qwen2-VL is processed by a domain-adapted Llama model, which employs inductive reasoning to produce expert-level fault interpretations and maintenance recommendations. All outputs are integrated into a structured fault detection and diagnosis report.

\subsection{Functional construction and support toolset for the system framework}
\label{subsec:sample2-1}
\begin{figure}[htbp]
    \centering
    \includegraphics[width=1\linewidth]{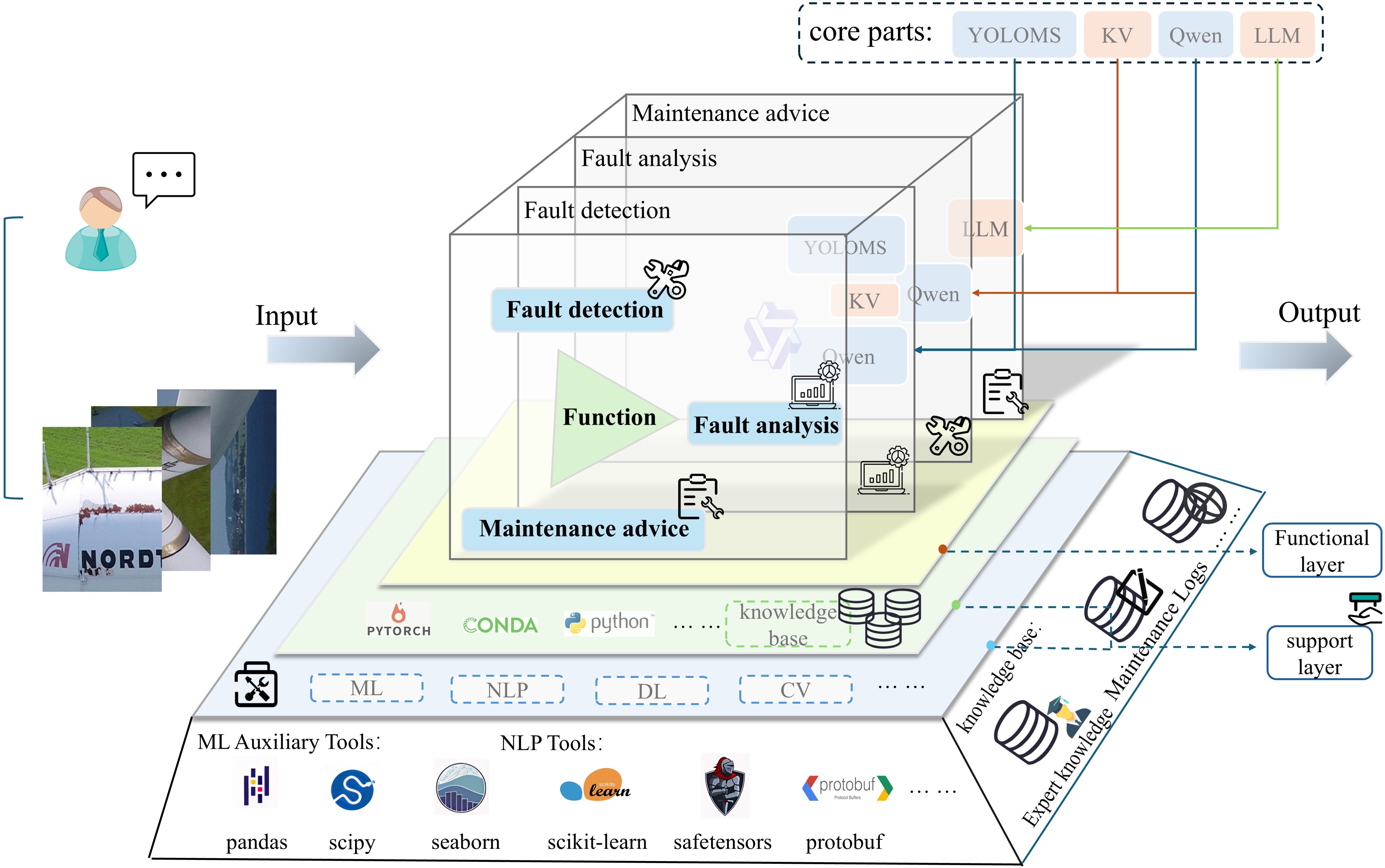}
    \caption{ LLM-assisted fault detection and diagnostic support and functional framework for WT components using the YOLOMS model.}
    \label{fig:2}
\end{figure}

The system consists of three core functional modules: fault detection, fault analysis and maintenance advice. Fault detection realizes anomaly identification, fault analysis analyzes the causes of faults in depth, and maintenance advice outputs maintenance solutions. The proposed framework support and functionality
 are shown in Fig.\ref{fig:2}. The core part integrates with YOLOMS, KV, Qwen2-VL and Llama, and operates in collaboration with the support layer through the function layer. The support layer covers ML, NLP, DL, CV and other technical fields, integrates ML auxiliary tools (pandas, scipy, seaborn), NLP tools (scikit-learn, safetensors, protobuf), and builds a knowledge base that contains expert knowledge, maintenance logs and other contents.The underlying architecture of this system is established on PyTorch, CONDA, and Python technologies. It relies on the knowledge base module to ensure the implementation of functions. A comprehensive processing flow from input to output is created. This flow is suitable to meet the application requirements for the fault detection, analysis, and maintenance of WT components in industrial scenarios.  

\subsection{The proposed YOLOMS model}
\label{subsec:sample2-2}
\begin{figure}[htbp]
    \centering
    \includegraphics[width=1\linewidth]{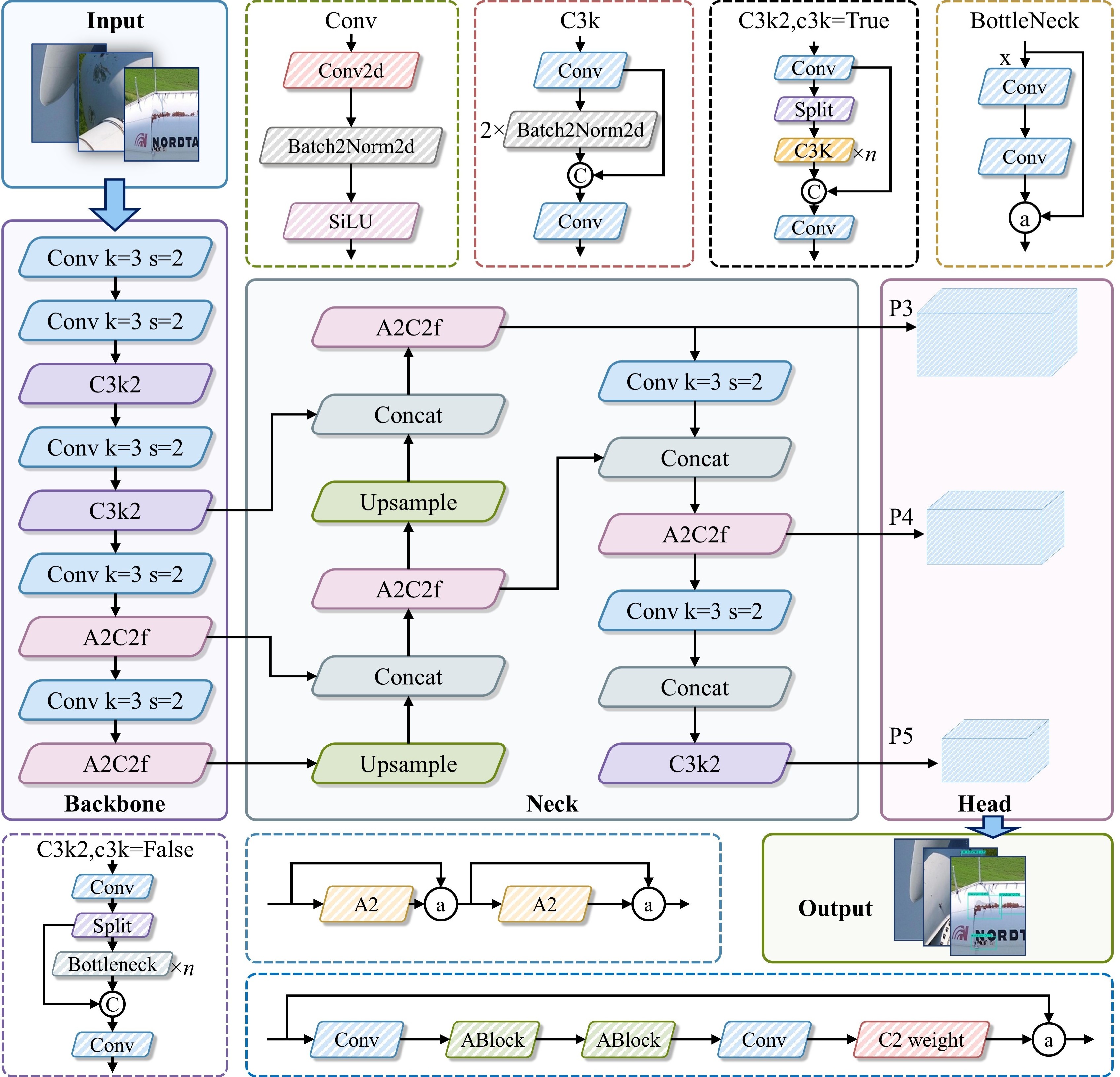}
    \caption{YOLOv12 network structure diagram.}
    \label{fig:3}
\end{figure}

YOLOv12 is the most recent version of the YOLO family, designed to enable faster inference and higher detection accuracy for a model that is jointly proposed in 2025 by a team of researchers from the University at Buffalo and the University of Chinese Academy of Sciences. Compared with other models in the YOLO series, YOLOv12 breaks the dominance of CNN models in the series and makes effective optimizations. For example, it uses FlashAttention and the regional attention module(A2) to enhance feature extraction, and proposes the Residual Efficient Layer Aggregation Network (R-ELAN) to optimize the efficiency of the attention mechanism. The Fig.\ref{fig:3} shows the network structure of YOLOv12.

Table \ref{tab:1} lists the YOLOv12 network structure for YAML-based configurations, covering six columns of key information. Layer represents the network layer number, which is used to pinpoint each layer. 'From' indicates the input source, such as '-1' or '[-1, 6]', which reflect different input scenarios; 'n' reflects the number of module repetitions, which improves the processing of the features. “module” contains various types, such as “Conv”, “Upsample”, “Concat”, “C3k2”, “Detect”, “A2C2f”, etc., each of which carries a unique function; “parameters” represents the number of parameters corresponding to the module, while “arguments” are the logic of the network flow is clear. At the beginning, the model relies on modules such as “Conv” to extract features. The Conv layer analyzes the input image layer by layer based on parameters such as convolution kernel and step size, obtains multi-scale and multi-dimensional features, and generates a feature map to lay the foundation for subsequent operations. Then, “Upsample” and “Concat” operations are applied to fuse the features. “Upsample” increases the size of the feature map, while “Concat” integrates different layers of features into a single feature to enrich the feature dimension. Finally, the “Detect” module locates the target and determines the category based on the fused features, and outputs reliable target detection results.

\begin{table}[H]
\centering
\caption{The network structure of YOLOMS.}
\label{tab:1}
\begin{tabular}{c c c c c l}
\toprule
\textbf{Layer} & \textbf{From} & \textbf{N} & \textbf{Params} & \textbf{Module} & \textbf{Arguments} \\
\midrule
0  & -1           & 1 & 464     & Conv     & {[}3, 16, 3, 2{]} \\
1  & -1           & 1 & 4672    & Conv     & {[}16, 32, 3, 2{]} \\
2  & -1           & 1 & 6640    & C3k2     & {[}32, 64, 1, False, 0.25{]} \\
3  & -1           & 1 & 36992   & Conv     & {[}64, 64, 3, 2{]} \\
4  & -1           & 1 & 26080   & C3k2     & {[}64, 128, 1, False, 0.25{]} \\
5  & -1           & 1 & 147712  & Conv     & {[}128, 128, 3, 2{]} \\
6  & -1           & 2 & 180864  & A2C2f    & {[}128, 128, 2, True, 4{]} \\
7  & -1           & 1 & 295424  & Conv     & {[}128, 256, 3, 2{]} \\
8  & -1           & 2 & 689408  & A2C2f    & {[}256, 256, 2, True, 1{]} \\
9  & -1           & 1 & 0       & Upsample & {[}None, 2, 'nearest'{]} \\
10 & {[}-1, 6{]}   & 1 & 0       & Concat   & {[}1{]} \\
11 & -1           & 1 & 869012  & A2C2f    & {[}384, 128, 1, False, -1{]} \\
12 & -1           & 1 & 0       & Upsample & {[}None, 2, 'nearest'{]} \\
13 & {[}-1, 4{]}   & 1 & 0       & Concat   & {[}1{]} \\
14 & -1           & 1 & 24000   & A2C2f    & {[}256, 64, 1, False, -1{]} \\
15 & -1           & 1 & 36992   & Conv     & {[}64, 64, 3, 2{]} \\
16 & {[}-1, 11{]}  & 1 & 0       & Concat   & {[}1{]} \\
17 & -1           & 1 & 74624   & A2C2f    & {[}192, 128, 1, False, -1{]} \\
18 & -1           & 1 & 147712  & Conv     & {[}128, 128, 3, 2{]} \\
19 & {[}-1, 8{]}   & 1 & 0       & Concat   & {[}1{]} \\
20 & -1           & 1 & 378880  & C3k2     & {[}384, 256, 1, True{]} \\
21 & {[}14, 17, 20{]} & 1 & 431452 & Detect   & {[}4, {[}64, 128, 256{]}{]} \\
\bottomrule
\end{tabular}
\end{table}

Due to the scarcity of WT component fault samples and poor detection performance, this paper proposes the innovative YOLOMS model. YOLOMS stands for YOLO Multi-Scale, which uses advanced techniques such as multi-scale cropping and sliding window cropping to enrich the dataset and effectively improve feature extraction efficiency. 

When a $W*H$ original image is fed into the YOLOMS module, it preprocesses it using multi-scale and sliding window cropping. The base window size is $W_B*H_B$, and $K$ windows of different sizes are generated using a scaling factor $r$, The size $W*H$ of the kth window is as follows:

Multiple windows of different sizes are generated using the above formula and cropped to obtain fault features at different resolutions. For example, a small window can focus on subtle localized faults, while a large window can capture large-scale defects. This addresses the issue of sample scarcity and improves the model's robustness to faults of varying sizes.
The sliding window technique further increases sample diversity through overlapping cropping. The sliding step size $s$ is determined by the overlap ratio $o$ and the window size:

\begin{equation}
\begin{cases}
W_k = W_B \times r^k \\
H_k = H_B \times r^k
\end{cases}
\label{eq:scale}
\end{equation}

Multiple windows of different sizes are generated using the above formula and cropped to obtain fault features at different resolutions. For example, a small window can focus on subtle localized faults, while a large window can capture large-scale defects. This addresses the issue of sample scarcity and improves the model's robustness to faults of varying sizes.

The sliding window technique further increases sample diversity through overlapping cropping. The sliding step size $s$ is determined by the overlap ratio $o$ and the window size:

\begin{equation}
s = [W_k \times o]
\label{eq:s}
\end{equation}

For a window of width $W_k$: Starting from one edge of the image, move $s$ pixels toward the other edge at a time until the entire image width is covered. The number of sliding steps $n$ satisfies:

\begin{equation}
n = \left\lfloor \frac{W - W_k}{s} \right\rfloor + 1
\label{eq:n}
\end{equation}

Through sliding windows and overlapping strategies, a large number of sub-images can be generated from the same original image, which not only increases the size of the dataset but also captures the characteristics of faults at different locations, enhancing the model's generalization ability for fault locations.

Based on this, YOLOv12 is used for detection. As a result, the model can accurately detect faults of different sizes and positions, even under challenging imaging conditions. This method significantly improves detection accuracy and demonstrates robustness and adaptability in real-world applications.

\subsection{Key-Value lightweight image-text mapping module}
\label{subsec:sample2-3}
Key-Value lightweight image text mapping is a data storage and retrieval method that links keys with graphic information through key-value pairs. The output of YOLO typically consists of images with recognition frames and corresponding labels, but overlapping tags can lead to information loss and ambiguity in data extraction. Moreover, directly inputting the original images and the YOLO output with recognition information into the LLM yields unsatisfactory textual feedback results. 

\begin{figure}[htbp]
    \centering
    \includegraphics[width=1\linewidth]{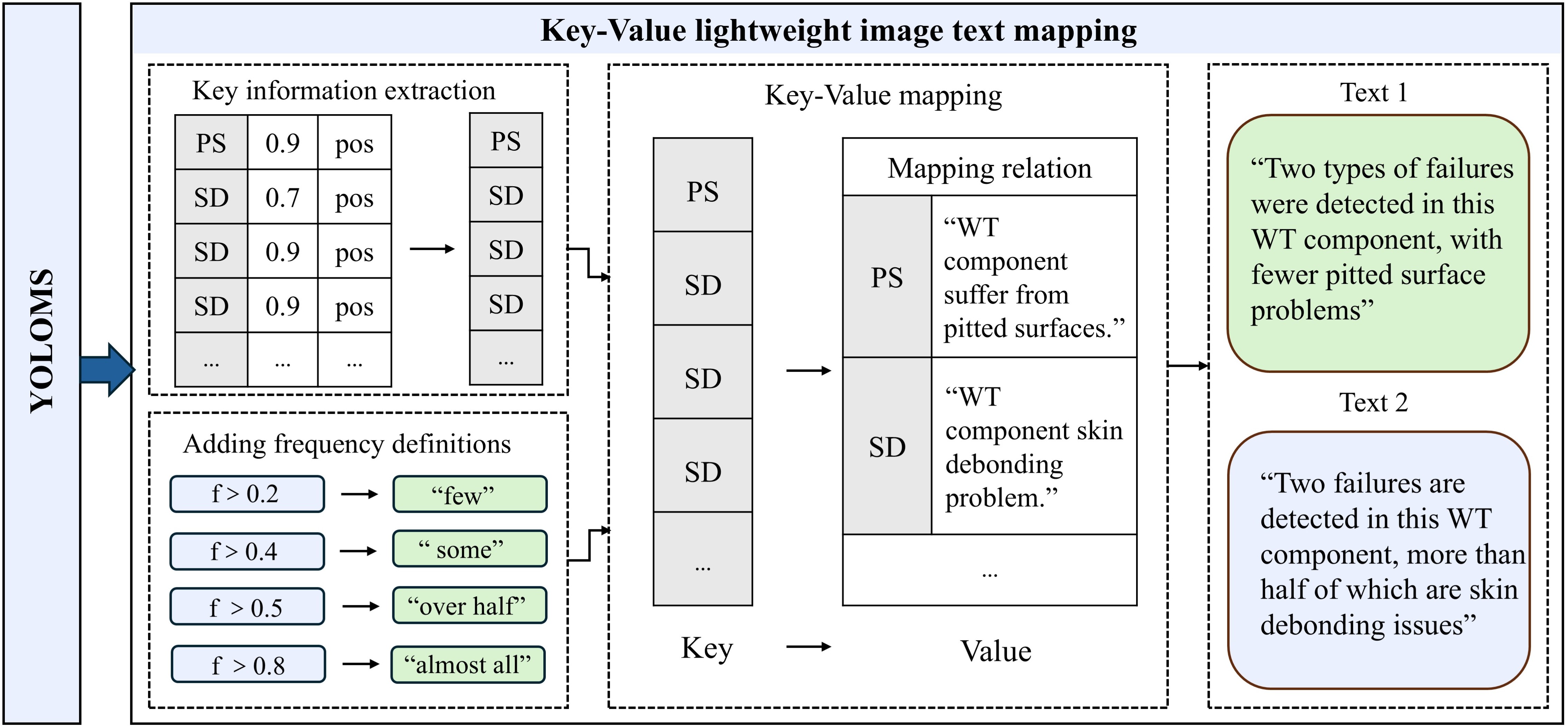}
    \caption{Structural diagram of the Key-Valued lightweight image-text mapping method. In the figure, “PS” and “SD” represent the abbreviations of the fault names, and “f” represents the frequency of occurrence of various faults.}
    \label{fig:4}
\end{figure}

To address the aforementioned issues, this study proposes a KV method. The multidimensional vectors output by the YOLO model are directly extracted, containing information such as fault labels, confidence, and location, thereby preventing the loss of labeling details. The information within these vectors are categorized into two groups: recognized fault categories and other attributes (including confidence, location, and other relevant data). The fault information categories are used as keys in the KV method, while the corresponding category texts serve as values. Each key is paired with its corresponding value, enabling the conversion of YOLO output images into text. The flow principle of this method is illustrated in Fig.\ref{fig:4}.

The Key-Value lightweight image-text mapping module is specifically designed to transform high-dimensional vectors output by YOLOMS into structured textual descriptions, functioning as a pivotal bridge between visual detection results and the interpretive capabilities of Large Language Models. It is important to clarify that the KV module eschews conventional dimensionality reduction techniques, such as feature selection or matrix decomposition. Instead, through semantic restructuring, it retains all critical information embedded in the original vectors, including fault labels, confidence scores, coordinates, and regional attributes, thereby mitigating information loss arising from label overlap during the conversion process.

To generate quantifiable textual descriptions, the KV module first calculates the occurrence frequency of each fault type. For a specific fault category $k$, let $N_k$ denote the number of detection instances, and $N_total$ denote the total number of all detected faults. The frequency $f_k$ is defined as eq.(\ref{eq:4}).

\begin{equation}
f_k = \frac{N_k}{N_{\text{total}}}
\label{eq:4}
\end{equation}

Based on $f_k$, descriptive quantifiers are assigned using thresholds calibrated via statistical analysis on the dataset, defined as:

\begin{equation}
\text{Quantifier}(f_k) = 
\begin{cases} 
\text{\lq few\rq} & 0.2 < f_k \leq 0.4 \\
\text{\lq some\rq} & 0.4 < f_k \leq 0.5 \\
\text{\lq over half\rq} & 0.5 < f_k \leq 0.8 \\
\text{\lq almost all\rq} & f_k > 0.8 
\end{cases}
\label{eq:quantifier}
\end{equation}

In addition, the graphical mapping method proposed in this study also takes into account the occurrence frequency of fault types and the area of the damaged region within the recognition bounding boxes. Based on the number of fault types detected by YOLOMS, the frequency of occurrence for each fault type is calculated, and a description with quantifiers is generated. Simultaneously, the area of the damaged region within the recognition bounding box is roughly estimated and compared, producing descriptions with area information. 

For each detected fault, the area of the damaged region within the bounding box is estimated using coordinate information output by YOLOMS. Given a bounding box with top-left corner $(x_1,y_1 )$ and bottom-right corner $(x_2,y_2 )$, the area $S$ is calculated as eq. (\ref{eq:6}).

\begin{equation}
|x_2 - x_1| \times |y_2 - y_1|
\label{eq:6}
\end{equation}

To determine if a fault is "large-area," a threshold $S_th$ is set. If $S>S_th$, the description "large-area" is appended to the corresponding fault type.

This mapping, which enriches semantic representation, ensures that textual outputs convey both qualitative and quantitative characteristics of fault distribution while also resulting in more accurate and concise descriptions capable of quickly and effectively conveying information. 

\begin{figure}[htbp]
    \centering
    \includegraphics[width=1\linewidth]{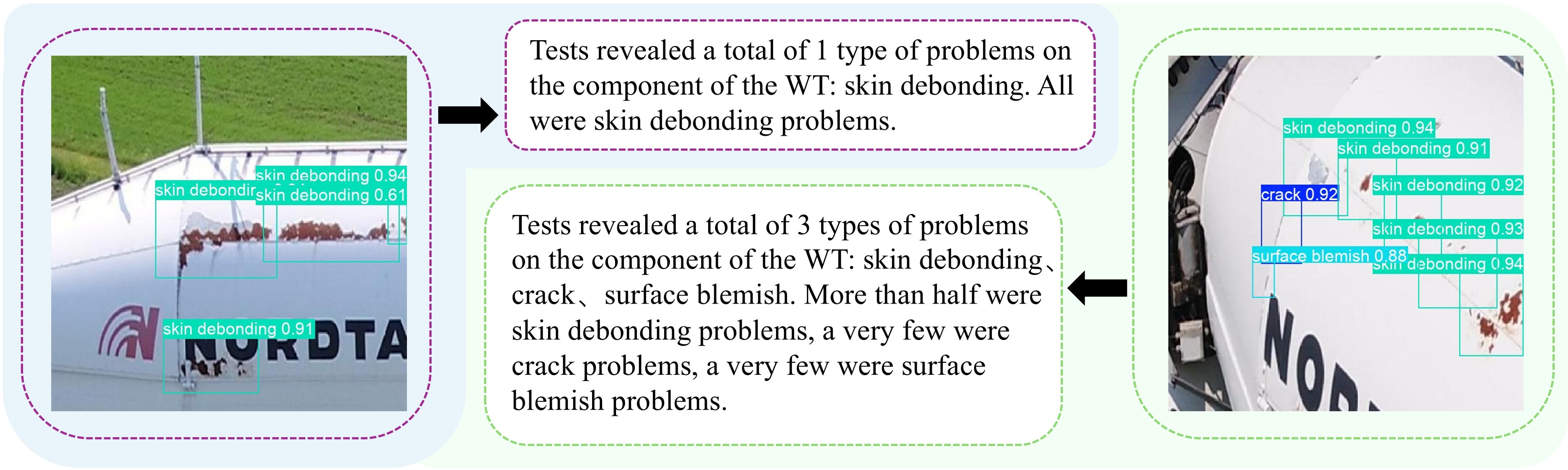}
    \caption{Two sets of graph mapping results, whose label vectors have been converted to text by the KV module.}
    \label{fig:5}
\end{figure}

As a result, it helps the LLM to better understand the fault types and their significance, providing a solid foundation for subsequent diagnostic analysis. The two sets of results mapped using this method are shown in Fig.\ref{fig:5}.

\subsection{LLM-assisted detection and diagnosis module}
\label{subsec:sample2-4}

To analyze WT component faults and provide actionable maintenance recommendations, this study integrates two key models: Qwen2-VL and Llama3.2, which play critical roles in enhancing fault detection and diagnosis.

Specifically, Qwen2-VL contributes significantly by leveraging its advanced multimodal capabilities, enabling YOLOMS to enhance the accuracy of fault identification and assessment. Through the fusion of textual descriptions and visual data, Qwen2-VL allows for a more comprehensive understanding of the fault types and their severity, providing an important foundation for more effective analysis. Meanwhile, Llama3.2 offers robust support for O\&M by utilizing its domain-specific finetuning to process the structured information from YOLOMS. The LLM’s role is not limited to interpreting image recognition results; it also employs deep reasoning based on expert knowledge, which leads to more precise fault analysis and maintenance recommendations. The integration of YOLOMS with the multimodal large model Qwen and the large language model Llama significantly improves the efficiency and accuracy of fault diagnosis, enabling the system to generate high-quality diagnostic reports in a much shorter time than traditional methods, ultimately helping to make more informed decisions for turbine maintenance.

\subsubsection{Qwen2-VL auxiliary detection and diagnosis}

\begin{figure}[h]
    \centering
    \includegraphics[width=1\linewidth]{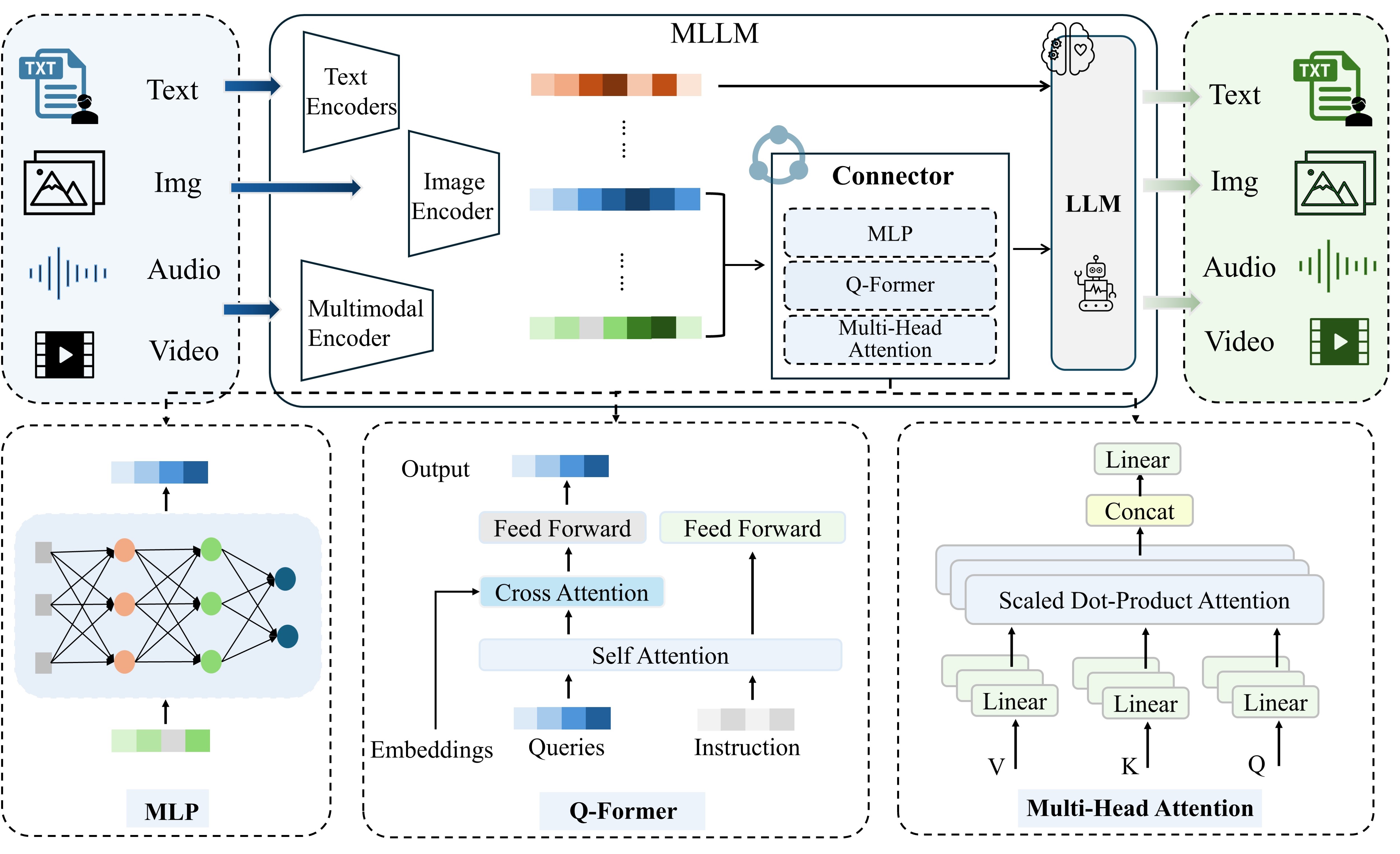}
    \caption{The system integrates multiple encoders to process heterogeneous data, which are fused with features by connectors and fed into a LLM for deep processing, and the results are output by a generator.}
    \label{fig:mllm}
\end{figure}

A multimodal large language model is an emerging type of large model that accepts inputs beyond just text, including images, audio, and video. By integrating information from different modalities, the model can better understand and generate content \cite{ma2025refinement,ding2024large}. The visualization of the functional process of multimodal large language models, including Qwen2-VL, is shown in Fig.\ref{fig:mllm}. After multimodal information is input, a modality encoder processes the data, extracts features, and converts the different modal inputs into a unified representation. The connector then integrates and aligns these features from different modality encoders for model processing. The integrated features are fed into the LLM, and the generator produces outputs such as text, images, audio, video, and other content.

There are three main components in the connector, which are Multi-Layer Perceptron (MLP), Query Transformer (Q-Former), and Multi-Head Attention. Among them, MLP is a feed-forward neural network consisting of multi-layer nodes, and the general form of the MLP can be expressed as eq. (\ref{eq:7}). 

\begin{equation}
\mathbf{h}^{(l+1)} = \sigma(\mathbf{W}^{(l)}\mathbf{h}^{(l)} + \mathbf{b}^{(l)})
\label{eq:7}
\end{equation}

Q-Former is based on the Transformer architecture, which can map the feature vectors of different modalities to a common feature space, thus realizing effective alignment and interaction between multiple modalities and improving the model’s ability to process complex information. Realizing the interaction between query vectors and embedding vectors can be achieved by calculating the correlation between the query and the embedding vectors, which is also known as Dot-Product Attention mechanism (DPA). Equation is shown in eq. (\ref{eq:8}).

\begin{equation}
\mathbf{A}_{ij} = \frac{\exp(\mathbf{q}_i^\top \mathbf{x}_j)}{\sum_{k=1}^n \exp(\mathbf{q}_i^\top \mathbf{x}_k)}
\label{eq:8}
\end{equation}

Multi-Head Attention is able to capture the dependencies between different locations in the input data, which can help the model to consider the feature information of different modalities at the same time, and improve the model’s comprehension and generation ability.

Qwen2-VL receives two inputs: WT component fault images and recognition frame diagrams processed by YOLOMS, and simple text descriptions generated by the KV method. Based on its pre-trained multimodal architecture, Qwen2-VL deeply integrates visual and textual information, accurately analyzes fault characteristics, and outputs key information covering fault types and severity, providing a solid foundation for subsequent maintenance decisions.

\subsubsection{Llama auxiliary operations and maintenance}

\begin{figure}[ht]
    \centering
    \includegraphics[width=1\linewidth]{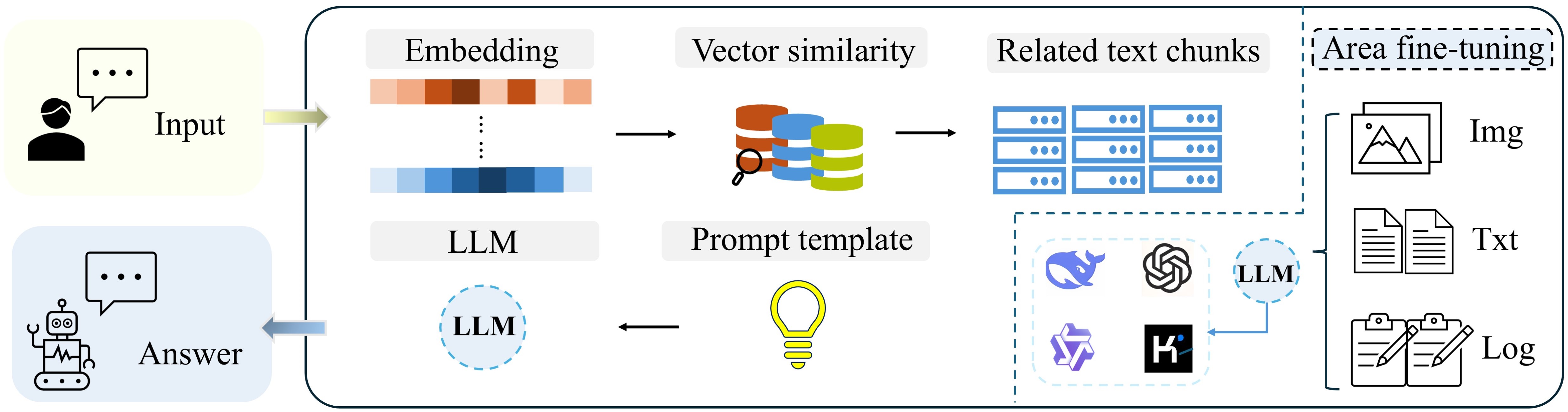}
    \caption{The main process can be broken down into five steps, with separate references in the right-hand section to the parts where LLM has fine-tuned its areas of specialization.}
    \label{fig:llm}
\end{figure}

Llama has powerful language processing capabilities, they learn patterns and structures of language by analyzing and understanding large amounts of textual data, and its flow is shown in Fig.\ref{fig:llm}. When text is input, it is first encoded within the model, converting natural language text into vector representations suitable for processing. The attention mechanism based on the Transformer architecture processes these vectors, focusing on key information. Next, the similarity between vectors is compared against a vector database, and the most relevant text blocks are retrieved. These relevant text blocks are then combined with a prompt template to form a new prompt. The enhanced prompt is fed into Llama, which generates responses based on its learned knowledge and patterns, and returns the output to the user. 

However, the original Llama model has several limitations when applied to tasks such as WT component fault detection and diagnosis. Since most of the data used during its pre-training comes from general domains, it lacks specialized knowledge specific to WT maintenance. As a result, when directly applied to WT fault detection and maintenance recommendations, its output often fails to meet the professional and accuracy requirements of real world operations. Therefore, fine tuning the Llama model with specialized knowledge in WT maintenance is essential. The model can better understand and process WT related information by deeply integrating professional knowledge from the WT maintenance process, summarizing actual expert maintenance reports and compiling a domain specific question-answer dataset.

For the domain-adaptive fine-tuning of the Llama model, the utilized dataset comprises 42 question-answer pairs related to wind turbine component faults, covering the causes, impacts, and maintenance strategies of four core fault types: cracks, skin debonding, surface blemish, and surface pitting. The data comes from multiple sources, such as actual operation and maintenance logs and the WT maintenance manual, and contains approximately 12,000 terms, averaging 280 terms per sample, including multi-dimensional information such as failure descriptions, causal analysis, and maintenance procedures. For cases where actual log coverage is insufficient, such as concurrent failures and extreme environment-triggered events, we generated supplementary samples using GPT based on core clauses of the wind turbine maintenance manual to cover rare failures and cross-failure correlation analysis.

To obtain a higher-quality fine-tuning dataset, this study performed unified cleaning on multi-source data. With reference to industry annotation manuals, steps such as deduplication, terminology standardization, and procedure format normalization were carried out to ensure that the dataset maintains structural consistency and complies with specifications in terms of content.  

In the fine-tuning process, the dataset was split into a training set and a validation set at an 8:2 ratio, ensuring that the training data covers all fault types with a distribution consistent with real operational scenarios. A full-parameter fine-tuning strategy was adopted, with a learning rate of 1e-4 and 5 training epochs. This enabled the Llama model to gradually acquire domain knowledge of wind turbines: actual operation and maintenance logs enhanced its understanding of real-world scenarios, while GPT-generated supplementary data improved its generalization ability for rare faults. Ultimately, the model achieved accurate associative reasoning across the "fault type-cause-maintenance strategy" chain.

After fine-tuning with WT maintenance knowledge, Llama takes the output text from the Qwen2-VL model as input and considers multiple factors to generate fault analysis and maintenance recommendations. In summary, the Llama module ultimately outputs a complete fault detection solution, including: fault type, fault severity, fault generation cause, and maintenance recommendation.

\section{Experimental results and comparative analysis}

The hardware configuration of the experiment is NVIDIA A100-80GB graphics card, CPU model is Intel(R) Xeon(R) Gold 6230R, and the training environment is Linux system. The python programming language is used with CUDA version 12.1.1 and PyTorch version 2.4.0.

\begin{figure}[H]
    \centering
    \includegraphics[width=1\linewidth]{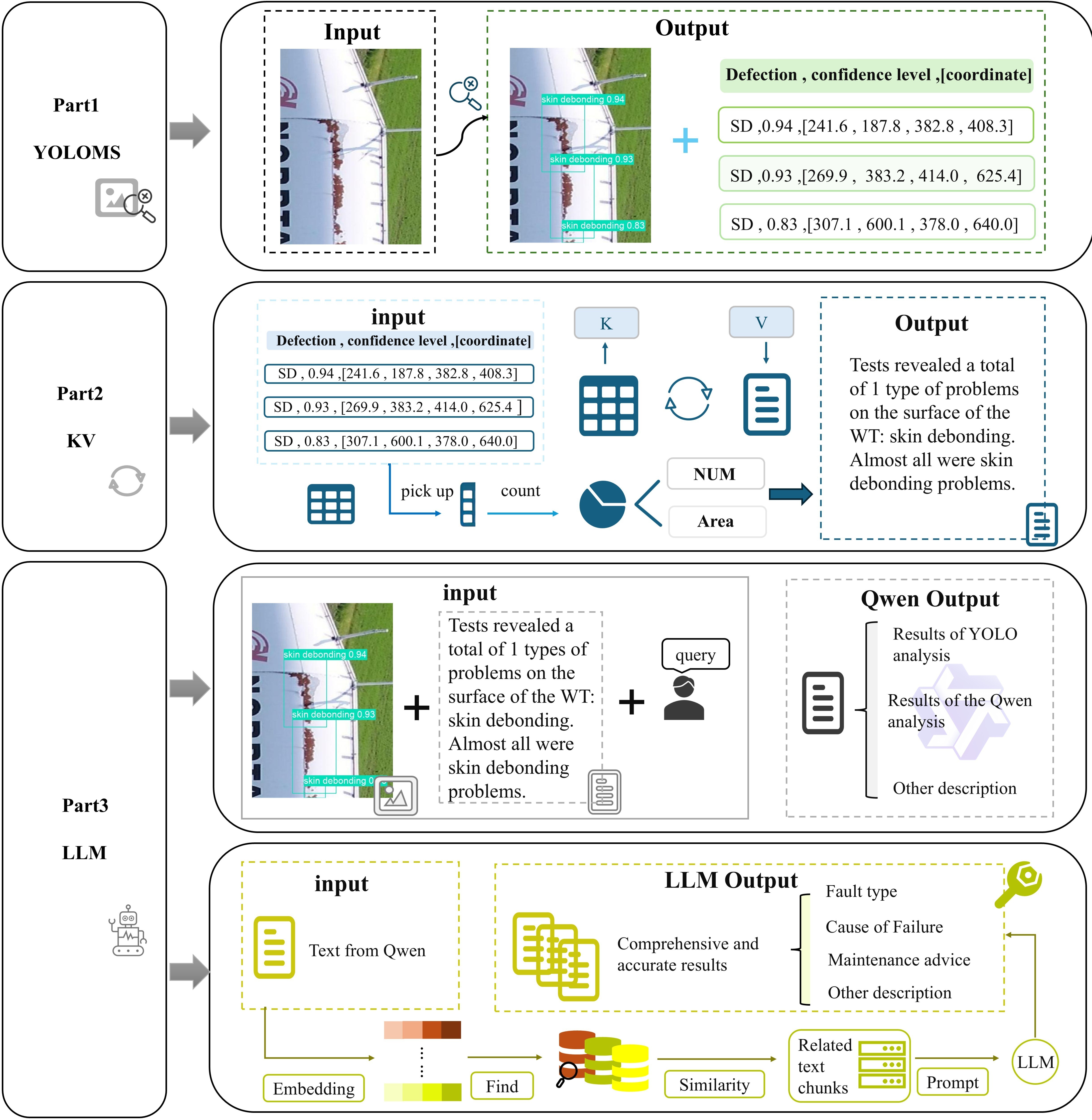}
    \caption{Input and output of each part of the framework.}
    \label{fig:8.}
\end{figure}

To help readers better understand the structure of the proposed framework clearly and intuitively, the entire framework has been divided into three parts based on the type of input. The details of the input and output elements for each part are shown in Fig.\ref{fig:8.}.

The first part is the YOLOMS section. This section takes the original WT image dataset as input. After the processing steps, it produces two types of output one is an image with detection boxes that clearly mark the key areas of interest in the image, and the other is a vector label containing information about the detection boxes, which records detailed characteristics of the detected regions in a structured format.

The second part is the KV section, which takes the vector information with detection box data output by the YOLOMS section as input. The KV section performs a specific transformation on these vectors and outputs concise and clear text descriptions. This makes it easier for the next part to process and analyze the information effectively.

The third part is the LLM section, which includes the previously mentioned Qwen2-VL and Llama modules. This section takes two types of input, the images output by the YOLOMS section and the text generated by the KV section. By comprehensively analyzing and processing both inputs, the LLM section ultimately produces complete and detailed detection results, providing strong support for subsequent decisions and operations.

\subsection{WT component actual measurement data}

The dataset utilized in this study was downloaded from the public dataset on the Kaggle platform. It was collected by a drone equipped with a high - resolution HD camera and comprises a total of 4048 images of WT. The WT component failure types are classified into four categories: cracks, surface blemish, skin debonding, and pitted surfaces. Images of components exhibiting these typical failure types are shown in Fig. 7, and the four failure types are detailed in Table \ref{tab:2}. The location and category of each defective area are labeled by experts, and these labels are used as the foundational data for analysis.

\begin{table}[ht]
\centering
\caption{failure types explanation}
\label{tab:2}
\begin{tabular}{@{}cc@{}}  
\toprule
\textbf{Defect categories} & \textbf{Explanation} \\
\midrule
Crack & Cracks appearing inside or on the surface of the component. \\
Skin debonding & Failure in bond between protective and structural layers. \\
Surface blemish & Slight localized superficial defects on the WT surface. \\
Pitted surface & Irregular depressions or pits in the surface. \\
\bottomrule
\end{tabular}
\end{table}

The size or number of possible faults on WT components usually varies widely, some damage may be a few centimeters while others may be larger. In addition, the complexity of the background during the shooting process and the conditions of distance, angle and lighting can affect the quality of the image inspection to varying degrees \cite{zou2022research,song2022review}.

To enhance the diversity and adaptability of image data, data augmentation strategies involving multi-scale cropping and sliding window cropping are employed during the data preprocessing stage. Specifically, original images are resized to a range of different resolutions in a differentiated manner, followed by cropping via the sliding window technique. A carefully selected set of cropping window dimensions and step intervals is adopted, and a well-designed overlap ratio is maintained throughout the sliding process to prevent missing critical defect features. This workflow yields a substantial quantity of sub-images with varied scales and spatial positions, which significantly improves the model’s performance in detecting defects of different sizes and locations. Furthermore, a validation and file replenishment mechanism is designed to supplement any missing images or labels, while also copying unprocessed original images along with their corresponding label files to the target directory to ensure the integrity and consistency of the dataset.

\begin{figure}[H]
    \centering
    \includegraphics[width=1\linewidth]{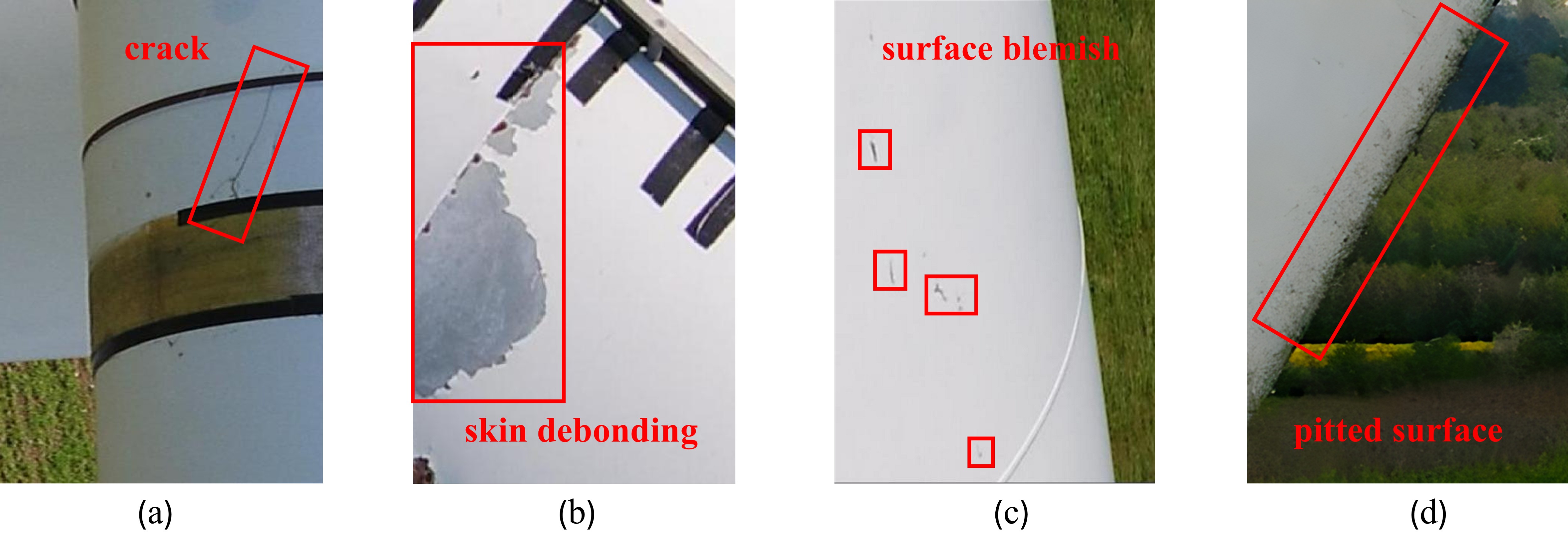}
    \caption{ Four failure types of WT components: (a) crack, (b) skin debonding, (c) surface blemish, (d) pitted surface.}
    \label{fig:placeholder}
\end{figure}

\subsection{YOLOMS model evaluation experiment}

To validate the advantages of the YOLOMS method proposed in this paper, fault detection experiments are conducted on WT components using several mainstream models, including YOLOv5, YOLOv8, YOLOv10, YOLOv11, and YOLOv12, under the same environment configuration and parameter settings. Each model is evaluated individually using performance metrics such as Accuracy (Acc), Precision (P), Mean Average Precision (mAP), Recall (Re), and F1 score (F1), as shown in eq. (\ref{eq:9}).

\begin{equation}
\begin{cases}
Acc = \dfrac{TP + TN}{TP + FP + FN + TN} \\
P = \dfrac{TP}{TP + FP} \\
mAP = \dfrac{1}{N}\sum_{i=1}^{N} AP_i \\
Re = \dfrac{TP}{TP + FN} \\
F1 = 2 \times \dfrac{P \times Re}{P + Re}
\end{cases}
\label{eq:9}
\end{equation}

Among them, the proposed YOLOMS model achieved the best results. Table3 lists the overall performance parameters and the mAP for the four failure categories (crack, skin peeling, surface blemish, and pitted surface) for all models in the experiment.

\begin{figure}[H]
    \centering
    \includegraphics[width=1\linewidth]{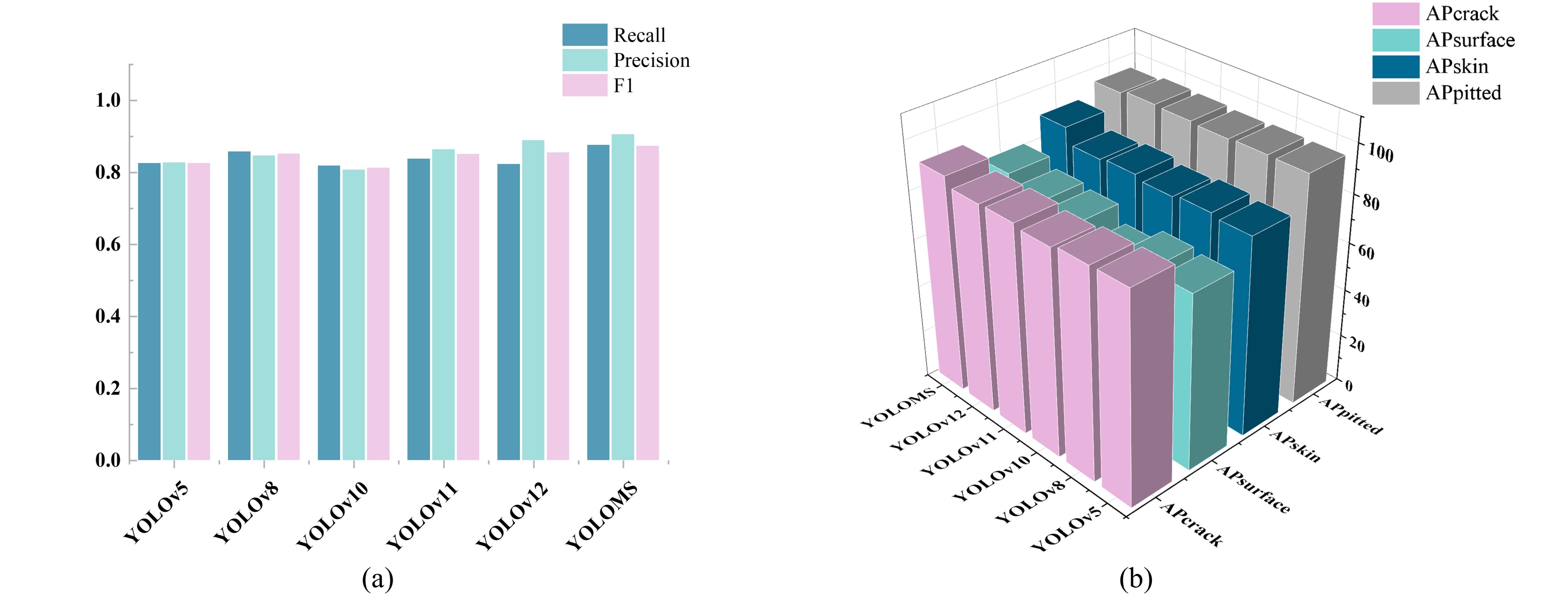}
    \caption{Data visualization of experimental results frommodels includingYOLOMS and YOLOv12.}
    \label{fig:YOLO}
\end{figure}

It is clear that the YOLOMS model performs best in all the metrics in the experiments, with its mAP50 metric being as high as 0.896, indicating that it has the best overall performance in the detection task. Its recall of 0.876 indicates that it performs best in recognizing positive samples. Its precision of 0.906 and F1 score of 0.873 show that it performs best in combining precision and recall. In addition, the Average Precision (AP) of all four categories of the YOLOMS model is above 0.8, and it has an APpitted of 0.955. The YOLOv8 and YOLOv12 models also perform well, with their mAP50 values of 0.857 and 0.864, respectively. These data highlight the superiority of the YOLOMS model in structure optimization and feature extraction, which enables it to recognize targets more efficiently in complex background environments. In order to make the information and trends in the data more intuitive and understandable to the readers, Fig.\ref{fig:YOLO} is the data visualization processing chart.

\begin{table}[ht]
\centering
\setlength{\tabcolsep}{0.8mm}  
\renewcommand{\arraystretch}{1.8}  
\caption{Comparison of experimental parameters}
\label{tab:exp_params}
\begin{tabular}{@{}ccccccc@{}}  
\toprule
\textbf{Parameters} & \textbf{YOLOv5} & \textbf{YOLOv8} & \textbf{YOLOv10} & \textbf{YOLOv11} & \textbf{YOLOv12} & \textbf{YOLOMS} \\
\midrule
mAP\textsubscript{50} & 0.850 & 0.857 & 0.837 & 0.863 & 0.864 & 0.896 \\
Recall                & 0.826 & 0.858 & 0.819 & 0.838 & 0.823 & 0.876 \\
Precision             & 0.827 & 0. 846& 0.807 & 0.864 & 0.889 & 0.906 \\
F1                    & 0.826 & 0.852 & 0.813 & 0.851 & 0.855 & 0.873 \\
AP\textsubscript{crack}  & 0.877 & 0.875 & 0.861 & 0.874 & 0.871 & 0.907 \\
AP\textsubscript{surface}  & 0.732 & 0.745 & 0.697 & 0.763 & 0.777 & 0.815 \\
AP\textsubscript{skin}   & 0.831 & 0.844 & 0.833 & 0.850 & 0.841 & 0.905 \\
AP\textsubscript{pitted} & 0.961 & 0.963 & 0.958 & 0.964 & 0.968 & 0.955 \\
\bottomrule
\end{tabular}
\end{table}
The proposed YOLOMS model is optimized on the basis of YOLOv12, and the comparison of the detection results between YOLOv12 and YOLOMS model outputs is shown in Fig. 8. It can be seen that the proposed model is more accurate in detecting faults in images of WT components and detects all types of labels with higher confidence.

After multi-scale cropping and sliding window cropping processing,the scale of the dataset increased significantly, and the confusion matrix is shown in Fig.\ref{fig:placeholder}(a). the number of images expanded from 3,543 to 36,520, and the total number of annotations rose from 6,055 to 69,680. Among them, the annotation counts for the four types of defects, including cracks and surface blemishes—increased by approximately 11.9 and 11.4 times, respectively. With balanced growth rates across all defect categories, this method effectively expanded the diversity of training samples and provided data support for the accurate detection of the YOLO model. The number of the remaining categories is also boosted, and the visualization image is shown in Fig.\ref{fig:placeholder}(b).

\begin{figure}[H]
    \centering
    \includegraphics[width=1\linewidth]{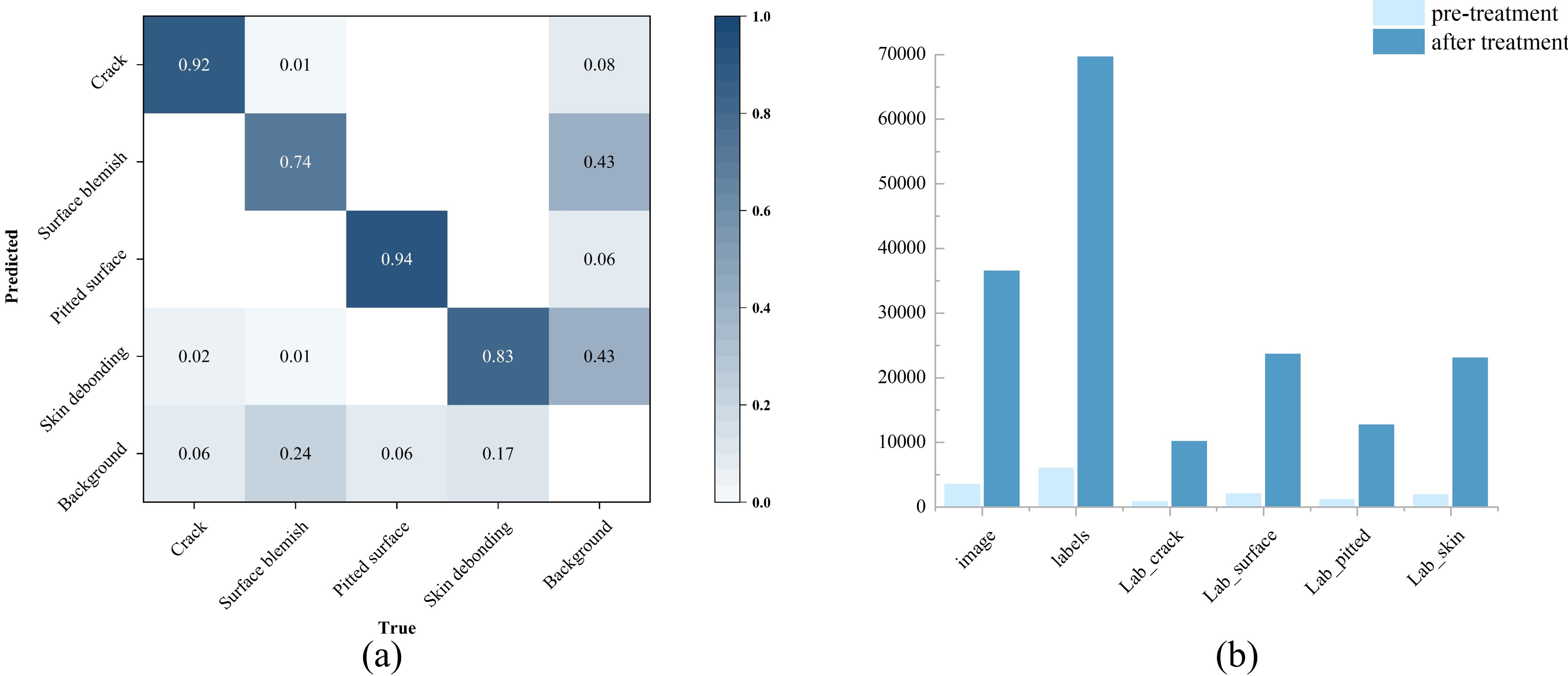}
    \caption{ (a) Confusion matrix of YOLOMS; (b) Comparison of the number of images and label data before and after processing.}
    \label{fig:placeholder}
\end{figure}

\subsection{LLM-assisted detection and diagnosis module evaluation experiment}
\label{subsec:sample3-3}

\subsubsection{Qwen2-VL auxiliary detection and diagnostic part evaluation experiment}
As the first stage of the input process in the LLM-assisted diagnostic module, Qwen2-VL essentially serves as an extension of the relevant test results. This means that its role goes beyond simply receiving information, more importantly, it performs further
analysis and processing based on the data obtained from the preceding tests, generating valuable textual outputs. In this context, the evaluation of the model focuses primarily on its graphical understanding capability. Graphical understanding refers to the model’s ability to accurately analyze the complex features, elements, and logical relationships embedded in the input images, which directly determines whether the generated text can comprehensively and accurately cover the fault categories detected by the YOLOMS module.

To scientifically, objectively, and effectively assess the model’s performance in this regard, after thorough consideration and comparative analysis, the Fault Consistency Score (FCS) is selected as the primary evaluation metric. In the calculation of FCS, a full-sampling method is employed. Full-sampling, as a rigorous and comprehensive data collection approach, ensures that the collected data samples cover all possible scenarios to the greatest extent, preventing inaccuracies caused by sample omissions or bias.

Fault Consistency Score: It is used to measure the extent to which the text generated by the Multi-modal Large Model covers the fault categories detected by YOLO. Let $C_y$  denote the total number of fault categories actually detected by the YOLO model, and $C_m$ denote the number of fault categories in the text generated by the large model that correctly match those detected by YOLO. Its calculation formula is expressed as eq. (\ref{eq:12}), namely:

\begin{equation}
FCS = \frac{C_m}{C_y}
\label{eq:12}
\end{equation}

After careful calculation and analysis of the large amount of data obtained from the full sampling, the average value is taken as the final evaluation result, and the average value is 0.94, which clearly shows that the outputs of Qwen2-VL and YOLOMS show
a high degree of consistency, which means that there is a good synergy and match between them in the identification and presentation of fault categories, which will provide a good basis for further analysis, decision-making, and application. This provides a solid and reliable data foundation and strong support for further analysis, decision-making and related applications.

\subsubsection{Llama auxiliary operations and maintenance part evaluation experiment}

In the proposed diagnostic system architecture, the fine-tuned Llama model, specialized in the relevant domain, plays a crucial role in the final stage of the process. As shown in Fig. 10, one of its core functions is to summarize the complex information generated by previous models. Leveraging advanced natural language processing techniques such as text analysis and semantic understanding, Llama identifies key points, establishes connections between different pieces of information, and presents the results in a clear and logically coherent manner. Additionally, Llama provides well-informed and actionable maintenance recommendations based on the summarized content, domain expertise, and extensive past cases. Through its complex reasoning and decision-making capabilities, it generates practical, scientific, and effective suggestions that guide subsequent maintenance efforts.

Given this critical role, the evaluation of Llama focuses on two key aspects, text quality and comprehensiveness. The output should be semantically clear, cover all essential diagnostic elements, omit no important details, and offer users a complete and reliable reference.

However, due to the limited availability of public diagnostic report data, selecting suitable evaluation metrics is essential to obtain accurate and objective results that truly reflect the model’s performance. After thorough analysis and comparison, this study selected the Average Precision Score (APS) as the evaluation metric for full-sample assessment. During APS evaluation, keywords are carefully chosen with professional expertise and a clear focus on text comprehensiveness and accuracy. These keywords reflect the core diagnostic content, highlight the key points of maintenance recommendations, and encompass various critical diagnostic elements. The APS is then calculated based on factors such as keyword occurrence, degree of match, and overall alignment with the generated text, providing an effective reference for measuring text quality and comprehensiveness.

Additionally, ablation experiments are conducted on several key modules, and the results are summarized in Table \ref{tab:5}. The final APS achieved by the proposed framework reached 0.89, demonstrating a high level of quality in terms of both content coverage and accuracy of presentation. This confirms that the framework delivers comprehensive and precise textual outputs, making it highly effective for real-world diagnostic and maintenance applications.

\begin{table}[ht]
\centering
\caption{Evaluation parameter results for three core module ablation experiments}
\label{tab:5}
\begin{tabularx}{\textwidth}{XXX>{\centering\arraybackslash}p{2cm}}  
\toprule
\multicolumn{3}{c}{\textbf{Method}} & \textbf{APS} \\  
\cmidrule{1-3}
\centering\textbf{YOLOMS} & \centering\textbf{Qwen2-VL} & \centering\textbf{Llama} &  \\  
\midrule
\centering$\checkmark$ & \centering$-$ & \centering$\checkmark$ & 0.74 \\
\centering$\checkmark$ & \centering$\checkmark$ & \centering$-$ & 0.71 \\
\centering$-$ & \centering$\checkmark$ & \centering$\checkmark$ & 0.67 \\
\centering$\checkmark$ & \centering$\checkmark$ & \centering$\checkmark$ & 0.89 \\
\bottomrule
\end{tabularx}
\end{table}

$\checkmark$ indicates modules to be considered for inclusion, while indicates modules not to be included. The information in the table shows that the proposed modeling framework gets the highest score, confirming the necessity of the existence of each component.

\begin{figure}[H]
    \centering
    \includegraphics[width=1\linewidth]{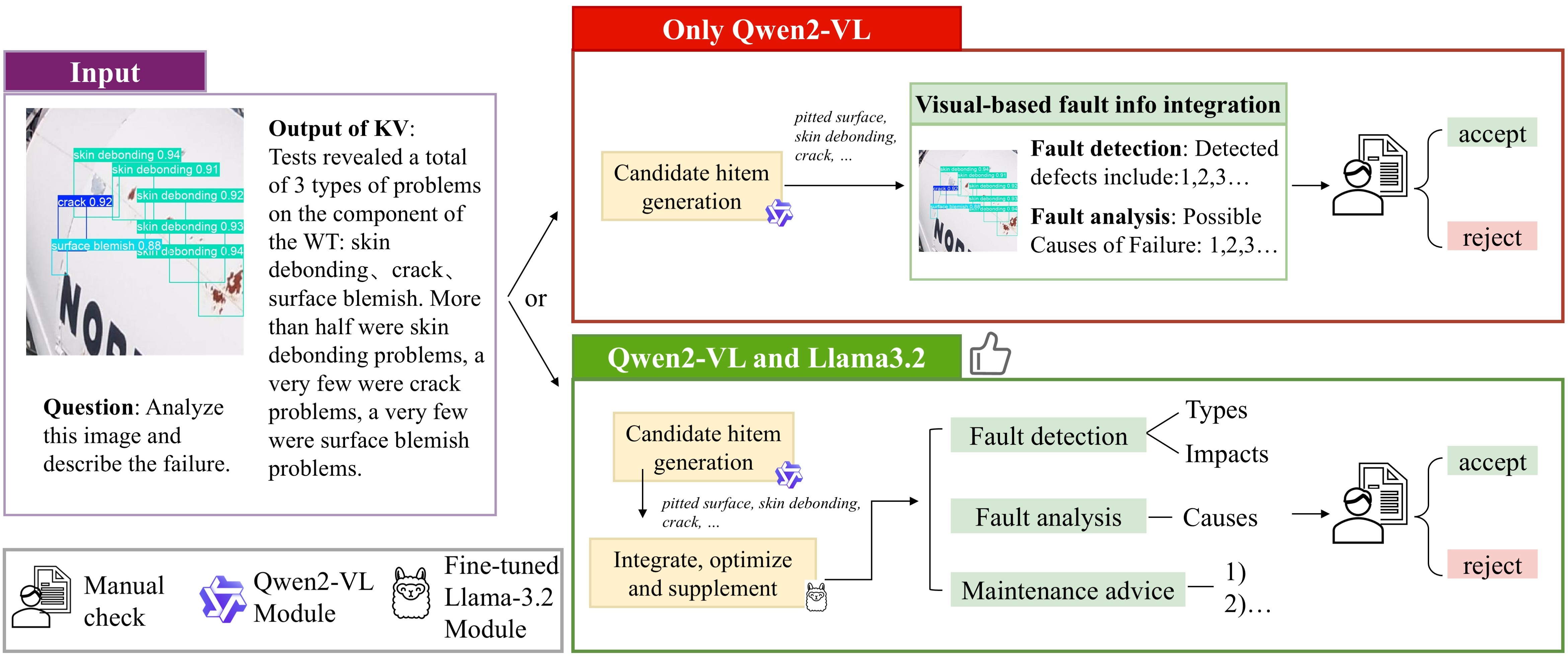}
    \caption{Comparison of output integrity of the large-model-assisted diagnosis module under different model combinations.}
    \label{fig:compare}
\end{figure}

In terms of the textual results output by the framework, this study further verifies the respective necessity of the Qwen2-VL model and the fine-tuned Llama collaborative diagnosis module. Fig.\ref{fig:compare}  compares the output integrity of the large-model-assisted diagnosis module under two scenarios: one is the scenario containing only the Qwen2-VL model, and the other is the scenario containing both the Qwen2-VL model and the fine-tuned Llama model. The outputs indicate that the model with the synergy of the two can yield better results. The complete textual output of the diagnostic framework is shown in Fig.\ref{fig:all} of the appendix.

The proposed framework can accomplish the functions of fault detection type, fault analysis and maintenance advice. The fault analysis content includes information such as the cause of the fault. In order to better enable the reader to understand the functions and outputs of each part, the text generated within the ablation experiment is summarized and outlined to make it clear which category the outputs of each part belong to as mentioned earlier. The summarized results are shown in Table \ref{tab:6}.

\begin{table}[ht]
\centering
\setlength{\tabcolsep}{1pt}  
\caption{Output category results for three core module ablation experiments}
\label{tab:6}
\resizebox{\linewidth}{!}{
\begin{tabular}{cccc}
\toprule
\multirow{2}{*}{\textbf{Method}} & \multicolumn{3}{c}{\textbf{Output}} \\
\cmidrule{2-4}
 & \textbf{Fault detection} & \textbf{Fault analysis} & \textbf{Maintenance advice} \\
\midrule
YOLOMS-Llama          & $\checkmark$ & $-$ & $\checkmark$ \\
YOLOMS-Qwen2-VL       & $\checkmark$ & $\checkmark$ & $-$ \\
Qwen2-VL-Llama        & $-$ & $\checkmark$ & $\checkmark$ \\
YOLOMS-Qwen2-VL-Llama & $\checkmark$ & $\checkmark$ & $\checkmark$ \\
\bottomrule
\end{tabular}
}
\end{table}

In summary, the complete structural framework containing three parts proposed in this study has the best comprehensive performance, fully utilizes the advantages of each module, and forms a more complete framework for WT component fault detection.

\section{Conclusions}
\label{sec:sample4}

The stability and reliability of wind turbines are closely related to the health of their components. Currently, wind turbine component fault detection faces two major challenges: first, the scarcity of fault samples leads to poor detection performance; second, structured output ignores semantic interpretability and lacks effective support for fault analysis and operation and maintenance strategy formulation. To address the problem of poor detection performance caused by scarce samples, this paper proposes an improved YOLOMS model. This model enriches the component dataset through multi-scale pruning and sliding window pruning techniques, thereby enhancing the fault feature extraction capability and detection performance. To address the problems of fault analysis and operation and maintenance strategy formulation, this paper combines the advantages of LLM and YOLO to propose the LLM-YOLOMS intelligent diagnostic framework. At the same time, a lightweight key-value (KV) mapping module is used to bridge the gap between visual output and text input, enabling it to be applied to fault diagnosis and analysis and provide maintenance suggestions. The study draws the following conclusions:

(1)	The proposed LLM-YOLOMS framework integrates multiple technological enhancements, combining the fault feature extraction capability of YOLOMS with the semantic reasoning strength of LLMs. Experimental evaluations show that YOLOMS achieves a fault detection accuracy of 90.6\%, while the generated diagnostic reports reach an average score of 89\% in textual quality and informational completeness. These results demonstrate that the framework provides reliable and comprehensive support for fault diagnosis and maintenance decision-making.

(2)	Within the framework, the YOLOMS model performs effectively in detecting faults across various wind turbine components. The mAP for crack detection reaches 90.7\%, surface defect detection 81.5\%, skin debonding detection 90.5\%, and pitted surface detection 95.5\%. These results confirm the model’s robustness and accuracy in identifying diverse fault types under real-world operational conditions.

(3)The proposed KV mapping method bridges the information gap between YOLOMS and the LLM by converting multidimensional detection vectors into structured textual descriptions, thereby preventing information loss caused by overlapping image labels. The method integrates both qualitative and quantitative fault characteristics to produce precise text representations, ensuring high-quality semantic inputs for subsequent fault analysis and diagnosis.

(4)The LLM achieves deep integration of visual and textual information, enabling comprehensive analysis of fault features. The domain-specific fine-tuned LLM demonstrates strong language understanding and generation capabilities, optimizing detection outputs, analyzing fault causes, and producing complete diagnostic reports that enhance engineers’ decision-making and maintenance planning.

\section*{CRediT authorship contribution statement}
\textbf{Yaru Li:} Data curation, Investigation, Methodology, Validation, Visualization, Writing-original draft. \textbf{Yanxue Wang:} Data curation, Methodology, Visualization, Funding acquisition, Supervision, Project administration, Writing-review \& editing. \textbf{Meng Li:} Supervision, Writing – review and editing. \textbf{Xinming Li:} Writing – review \& editing, Resources, Conceptualization.\textbf{Jianbo Feng:} Software, Supervision, Visualization.

\section*{Declaration of competing interest}
The authors declare that they have no known competing financial interests or personal relationships that could have appeared to influence the work reported in this paper.

\section*{Data availability}
Data will be made available on request.

\section*{Declaration of generative AI and AI-assisted technologies in the writing process}
During the preparation of this work the authors used ChatGPT-4o in order to improve readability and language of this work. After using this tool/service, the authors reviewed and edited the content as needed and take full responsibility for the content of the publication.

\section*{Acknowledgments}
This study was supported by the National Natural Science Foundation of China (52275079), the Youth Beijing Scholars program, the Guangxi Science and Technology Major Project (AA23062031).

\appendix
\section{LLM output including maintenance recommendations}
\begin{figure}[H]
    \centering
    \includegraphics[width=1\linewidth]{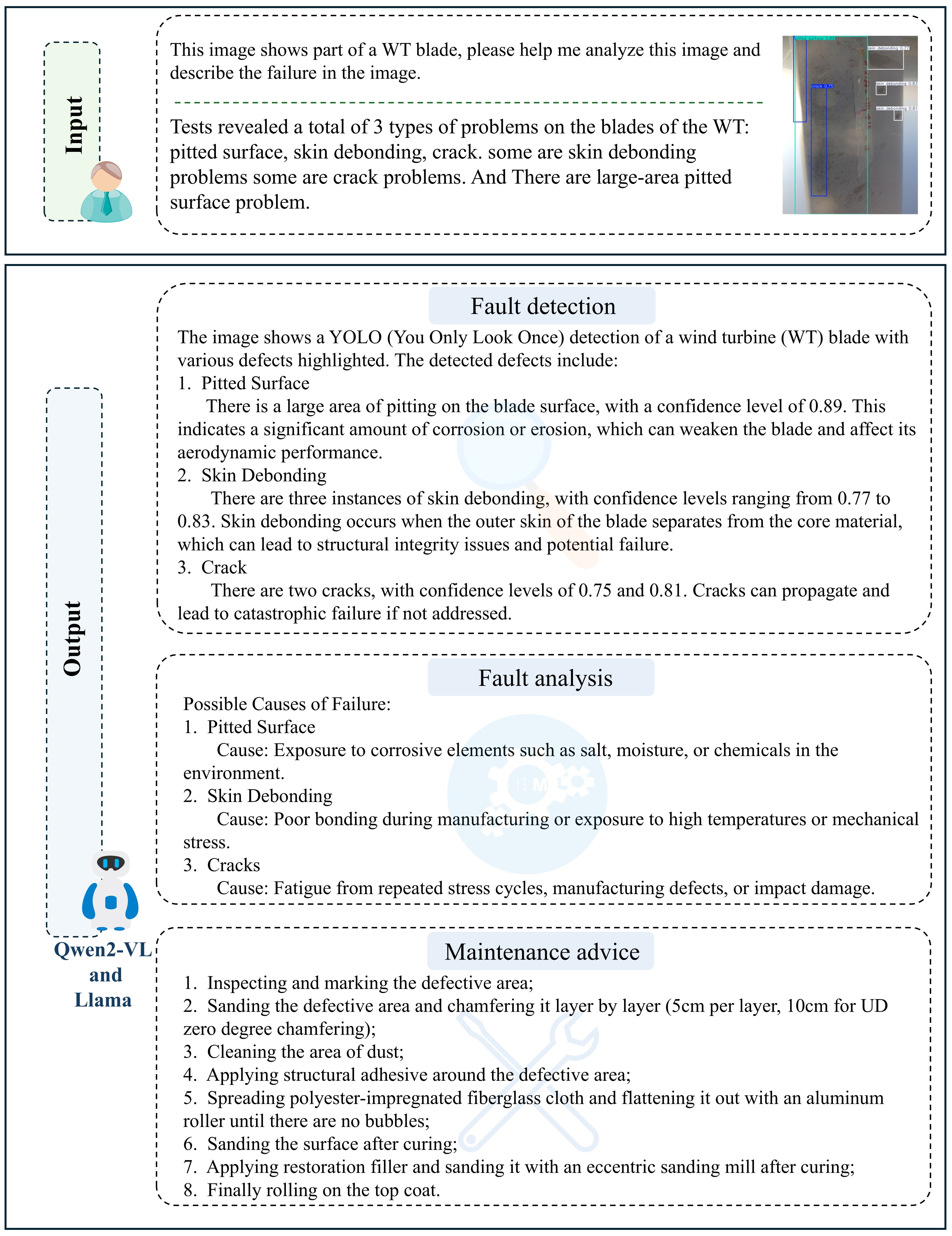}
    \caption{Example Dialog Box of Complete Text Output for the WT Blade Fault Detection Framework.}
    \label{fig:all}
\end{figure}



\renewcommand*{\bibfont}{\small}
\bibliographystyle{elsarticle-num} 
\bibliography{cas-refs}





\end{document}